\documentclass[conference]{IEEEtran}
\IEEEoverridecommandlockouts
\usepackage{graphicx}

\usepackage{comment}

\usepackage{bbm}
\usepackage{bm}

\usepackage{pgfplots}
\pgfplotsset{compat=1.7}
\usepackage[utf8]{inputenc}
\usepackage[english]{babel}

\usepackage{amssymb}   
\usepackage{stfloats}

\usepackage{mathtools}

\usepackage{accents}

 \usepackage{booktabs} 
\usepackage{cite}

\newcommand\y{\cellcolor{green!10}}

\newcommand\xA{.04}   
\newcommand\xB{0.18}  
\newcommand\xC{0.35}   
\newcommand\xD{0.7}   

 \usepackage{amsmath}
\usepackage{colortbl}
	\definecolor{arsenic}{rgb}{0.23, 0.27, 0.29}
	\definecolor{amaranth}{rgb}{0.9, 0.17, 0.31}
	\definecolor{aqua}{rgb}{0.0, 1.0, 1.0}
 
\usepackage{colortbl}
	\definecolor{arsenic}{rgb}{0.23, 0.27, 0.29}
	\definecolor{amaranth}{rgb}{0.9, 0.17, 0.31}
	\definecolor{aqua}{rgb}{0.0, 1.0, 1.0}

\newtheorem{theorem}{Theorem}

\newtheorem{proposition}{Proposition}

\newcommand{\taskRV}{\ensuremath{T}}
\newcommand{\task}{\ensuremath{\tau}}
\newcommand{\lenZm}{\ensuremath{\mathrm{M}}} 
\newcommand{  \lenDataSet}{\ensuremath{\mathrm{N}}} 
\newcommand{  \supSamp}{\ensuremath{\tilde{\bm Z}^{2\lenZm}_{1:2 \lenDataSet}}} 
\newcommand{  \supsamp}{\ensuremath{ {\bm y}^{2 \lenDataSet}}} 
\newcommand{  \supSamTask}{\ensuremath{\tilde{\bm Z}^{2\lenZm}_{1:2\lenDataSet}(\bm R)}}
\newcommand{\smZ}{\scriptstyle Z}
 
\newcommand{\metaGenLoss}{\ensuremath{\mathrm L(u)}}
\newcommand{\metaGenLossRV}{\ensuremath{\mathrm L(U)}}   
\newcommand{\Lhat}{\ensuremath{\widehat{\mathrm L}}}
\newcommand{\bmM}{ \tilde{\bm Z}^{2\lenZm}_{1:2\lenDataSet}}
\newcommand{\bmMT}{\tilde{\bm Z}^{2\lenZm}}
\newcommand{\genL}{\ensuremath{\widehat{\Delta \mathrm L}}}

\newcommand{\genLU}{\ensuremath{\overset{\sim}{\Delta\mathrm L}}}

\tikzset{every picture/.style={font issue=\scriptsize, >=stealth},font issue/.style={execute at begin picture={#1\selectfont}}}
    
\begin{document}

\title{Conditional Mutual Information-Based Generalization  Bound for Meta Learning\\
}
\author{\IEEEauthorblockN { Arezou Rezazadeh$^1$, Sharu Theresa Jose$^2$, Giuseppe Durisi$^1$, Osvaldo Simeone$^{2}$}\\
\IEEEauthorblockA { $^1$ Chalmers University of Technology, $^2$  Kings College London\\
{\small arezour@chalmers.se,sharu.jose@kcl.ac.uk,
durisi@chalmers.se,osvaldo.simeone@kcl.ac.uk
}
}
\thanks{This work has been funded by the European Union’s Horizon 2020
research and innovation programme under the Marie Sklodowska-Curie grant agreement No 893082 and  by  the  Wallenberg  AI,  autonomous  systems,  and software program.  It was also supported by the European Research Council (ERC) under the European Union’s Horizon 2020 Research and Innovation Programme (Grant Agreement No. 725731).
}
}

\maketitle

\begin{abstract}
Meta-learning optimizes an inductive bias---typically in the form of the hyperparameters of a base-learning algorithm---by observing
data from a finite number of related tasks.
This paper presents an information-theoretic  bound on the generalization performance of any given meta-learner, which builds on the conditional mutual information (CMI) framework of Steinke and Zakynthinou (2020).
In the proposed extension to  meta-learning, the CMI bound involves a training \textit{meta-supersample} obtained by first sampling $2\lenDataSet$ independent tasks from the task environment, and then drawing  $2\lenZm$  independent training samples for each sampled task.
The meta-training data fed to the meta-learner is modelled as being  obtained by randomly selecting $\lenDataSet$  tasks from the available $2\lenDataSet$ tasks and
$\lenZm$ training samples per task from the available $2\lenZm$ training samples per task. 
The resulting bound is explicit in two CMI terms, which measure
the information that the meta-learner output and the base-learner output  provide about which training data are selected, given the entire meta-supersample.
Finally, we present a numerical example that illustrates the merits of the proposed bound in comparison to prior information-theoretic bounds for meta-learning.
\end{abstract}

\section{Introduction}
Meta-learning refers to the process of automatically optimizing the hyperparameters of a training algorithm by observing data from a number of related tasks, so as to ``speed up'' the learning of a new, previously unseen task \cite{schmidhuber1987evolutionary, thrun1998learning}. Hyperparameters include the initialization and the learning rate of a training algorithm
\cite{finn2017model}, \cite{li2017meta}. As in prior information-theoretic analysis of learning systems \cite{Xu2017,Russo2016ControllingTheory,Bu}, we fix a 
training algorithm, also referred to as the \textit{base-learner}, as a stochastic mapping $\mathrm P_{W|\bm Z^\lenZm,U=u}$ from the input training data $\bm Z^\lenZm$ to the space of model parameters $\mathcal{W}$ for a given hyperparameter vector $u$. The meta-learner observes a meta-training data set $ \bm Z^\lenZm_{1:\lenDataSet}$, comprising of $\lenZm$ data samples, each from one of  $\lenDataSet$ related tasks, to optimize the hyperparameter $u$. The tasks are assumed to belong to a \textit{task environment}, which is defined by a task distribution $\mathrm P_\taskRV$ over a set $\mathcal{T}$ of tasks and by per-task data distributions $\{\mathrm P_{Z|\taskRV=\task}\}_{\task \in \mathcal{T}}$.

The goal of the meta-learner is to minimize the \textit{meta-generalization loss} $\metaGenLoss$, i.e., the average 
loss incurred by the hyperparameter~$u$ when used by the base-learner on a new task $\taskRV$ drawn  from the same task environment. 
However, this quantity is not computable since the task-environment distribution is unknown.
Instead, the meta-learner can evaluate the empirical \textit{meta-training loss} $\mathrm L_{\bm Z^\lenZm_{1:\lenDataSet}}(u)$ for hyperparameter $u$ based on the meta-training set $\bm Z^\lenZm_{1:\lenDataSet}
$. 
%
The difference between the meta-generalization loss and the meta-training loss, $\Delta\mathrm L(u|\bm Z_{1:\lenDataSet}^\lenZm)=\mathrm L(u)-\mathrm L_{{\bm Z_{1:\lenDataSet}^\lenZm}}(u)$, is the \textit{meta-generalization gap}. 
If the meta-generalization gap is small, 
the performance of the meta-learner on the meta-training set can be taken as a reliable measure of the meta-generalization loss. 
As in \cite{Russo2016ControllingTheory,Xu2017,Bu}, this paper studies information-theoretic bounds on the average meta-generalization gap, $\Delta \mathrm L^{\textup{avg}}=\mathbbm E_{\mathrm P_{U\bm Z^\lenZm_{1:\lenDataSet}}}  \left[\Delta\mathrm L (U|\bm Z_{1:\lenDataSet}^\lenZm)\right]$, where the average is taken over the meta-training set and the hyperparameters.
 
 \paragraph*{Related Work}
Following the initial work of Russo and Zhou~\cite{Russo2016ControllingTheory}, information-theoretic bounds on the average generalization gap for conventional learning
have been widely investigated in recent years~\cite{Xu2017,Bu,Negrea2019Information-TheoreticEstimates}. While the bound in~\cite{Xu2017} depends on the mutual information (MI) between the output of the learning algorithm and the training set, Bu \emph{et al.} \cite{Bu} tightened the bound via the individual sample mutual information (ISMI) between the algorithm output and each individual sample of the training set.
The MI and ISMI-based approaches  have been  extended to meta-learning in~\cite{sharu}.

Directly relevant to this paper is the approach 
recently 
introduced in~\cite{Steinke2020} for conventional learning, which  will be referred as \emph{conditional mutual information} (CMI) framework. Differently from MI-based bounds, the CMI-based bound given in~\cite[ Thm. 2]{Steinke2020} is always bounded. 
This is because the bound depends on the MI between the trained model and discrete data selection indices, rather than between trained model and  training data as in \cite{Steinke2020}.
To this end, the CMI framework introduces 
 a \emph{supersample} of $2\lenZm$ samples generated independently according to the underlying data distribution, and the training samples are chosen by selecting $\lenZm$ samples from the supersample at random. 
The resulting bound \cite[Thm. 2]{Steinke2020} depends on the CMI between the algorithm output and the random vector that determines which training data are selected from the supersample, conditioned on the supersample.

\paragraph*{Contributions}
Inspired by~\cite{Steinke2020}, we present  a novel information-theoretic bound on the average meta-generalization gap that extends the CMI-based bounds for conventional learning to meta-learning. 
In line with~\cite{sharu}, we show that there are two sources of generalization errors that contribute to the meta-generalization gap: (\emph{i}) the \emph{environment-level} meta-generalization gap, resulting from the observation of a finite number $\lenDataSet$ of tasks; and (\emph{ii}) the \emph{task-level} generalization gap, resulting from the observation of a finite number $\lenZm$ of data samples per task. Unlike prior work \cite{sharu}, these two contributions are quantified in our analysis via  CMI terms that depend on a \emph{meta-supersample} of per-task training data sets 
and on data selection indices
for  meta-learner and base-learners.
The derived bound inherits the advantage of the CMI bound for conventional learning, including its boundedness.  
We finally demonstrate the usefulness of the proposed bound on an example.

We conclude this section by further clarifying  the relation to previous works \cite{baxter, pentina2014pac, rothfuss2020pacoh, amit2018meta,denevi2018incremental,paper1,denevi2020advantage,paper4,finn2017model,paper3,paper2,paper5}.
The information-theoretic bounds on the average meta-generalization gap
derived here and in \cite{sharu} hold for arbitrary, fixed base-learners and meta-learners.
{
 In contrast, the probably-approximately-correct (PAC) bound of \cite{baxter} holds uniformly over a class of base-learners and meta-learners~\cite{zhang16-11a}. 
Our bound is related to,  although not directly comparable with,  the high-probability PAC-Bayes bounds reported in \cite{pentina2014pac,rothfuss2020pacoh,amit2018meta}, which also apply to arbitrary, fixed, base-learners and meta-learners.
}


References \cite{denevi2018incremental,paper1,denevi2020advantage,paper4,finn2017model,paper3,paper2,paper5} derive bounds on the optimality gap, $\mathbbm E_{\mathrm P_{U|\bm Z^\lenZm_{1:\lenDataSet}}}[\mathrm{L}(U)]-\min_{u \in \mathcal{U}}\mathrm{L}(u)$, on average or with high probability,
for  specific meta-learning instances and algorithms,  such as ridge regression with meta-learned bias in \cite{denevi2018incremental,paper1,denevi2020advantage,paper4},  
gradient based meta-learning methods in \cite{finn2017model,paper3}, and online meta learning 
in \cite{paper2,paper5}. 
In contrast, our derived bounds provide general information-theoretic insight into the number of tasks and number of samples per task required to ensure that the average meta-generalization gap for arbitrary base-learners and meta-learners is sufficiently small.

\section{Problem Definition}\label{sec:problemdefinition}
In this section, we provide the key definitions for our setup.
\paragraph*{Base-learner}
 Each task $\task$ within some set of tasks $\mathcal T$ is associated with an underlying unknown data distribution $\mathrm P_{Z|\taskRV=\task}$ on a set $\mathcal{Z}$. Throughout, sets can be discrete or continuous.
 For a given task $\task_i$, the \textit{base-learner} observes a data set $\bm Z^\lenZm_i=(Z^1_i,\dots,Z^\lenZm_i)$ of $\lenZm$ independently and identically distributed (i.i.d.) samples from 
 $\mathrm P_{Z|\taskRV=\task_i}$. 
The base-learner uses the training set $\bm Z^\lenZm_i$ to infer a model parameter $w\in\mathcal{W}$. 
We assume that the base-learner depends on an inductive bias that is defined by a hyperparameter vector $u\in \mathcal U$. 
The performance of the model parameter $w$ on a data sample $z$ is measured by the loss function $\ell : \mathcal W\times \mathcal Z\rightarrow \mathbbm R^+$. 
The goal of the base-learner is to infer the model parameter $w \in \mathcal{W}$ that minimizes the per-task generalization loss 
\begin{align}
    \mathrm L_{\mathrm P_{Z|\taskRV=\task_i}}(w)=\mathbbm E_{\mathrm P_{Z|\taskRV=\task_i}} [\ell(w,Z)],
\end{align} 
where the average is taken over a test sample $Z \sim \mathrm P_{Z|\taskRV=\task_i}$ drawn independently from $ \bm Z^\lenZm_i$. Since $\mathrm P_{Z|\taskRV}$ is unknown, the generalization loss  $\mathrm L_{\mathrm P_{Z|\taskRV=\task_i}}(w)$ cannot be computed. 
Instead, the base-learner evaluates the training loss
\begin{align}
    \mathrm L_{\bm Z^{\lenZm}_i}(w)=\frac{1}{\lenZm}\sum_{j=1}^\lenZm \ell(w,Z_i^j).\label{emp_loss_ordinary}
\end{align} 
The difference between the generalization loss and the training loss is referred to as the \emph{generalization gap}:
\begin{align}
   \Delta \mathrm L(w|\bm Z^{\lenZm}_i,u,\task_i)=\mathrm L_{\mathrm P_{Z|\taskRV=\task_i}}(w)-  \mathrm L_{\bm Z^{\lenZm}_i}(w).
       \label{2020.06.22_23.18}
\end{align} 
In this paper, we model the base-learner as a stochastic map $\mathrm P_{W|\bm Z^\lenZm_i, U=u}$ from the input training data set $\bm Z^\lenZm_i$ to the model class $\mathcal{W}$. As mentioned, this map depends on 
$u$.

\paragraph*{Meta-Learner} 
The goal of  meta-learning is to automatically infer the hyperparameter $u$ of the base-learner $\mathrm P_{W| \bm Z^\lenZm,U=u}$ from training data pertaining a number of related tasks. 
The tasks are assumed to belong to a \emph{task environment}, which is defined by a task distribution $\mathrm P_{\taskRV}$ on  the space of tasks $\mathcal T$, and by the per-task data distributions $\{\mathrm P_{Z|\taskRV=\task}\}_{\task \in \mathcal{T}}$.
The meta-learner observes a meta-training set $\bm Z_{1:\lenDataSet}^\lenZm=(\bm Z_1^\lenZm,\dots,\bm Z_\lenDataSet^\lenZm)$ of $\lenDataSet$ data sets. 
Each $\bm Z_i^\lenZm$ is generated independently by first drawing a task $\taskRV_i\sim \mathrm P_\taskRV$ and then a task-specific dataset  $\bm Z_i^\lenZm\sim \mathrm P_{\bm Z^\lenZm|\taskRV_i}$.

The meta-learner uses the meta-training set $\bm Z_{1:\lenDataSet}^\lenZm$ to infer the hyperparameter $u$.  
This is done with the goal of ensuring that, using the inferred hyperparameter $u$, the base-learner $\mathrm P_{W|\bm Z^\lenZm,U=u}$ can efficiently learn on a new {meta-test task} $ \taskRV \sim \mathrm P_\taskRV$ given the corresponding training dataset $\bm Z^\lenZm$. 

Formally, the objective of the meta-learner is to infer the hyperparameter $u$ that minimizes the \textit{meta-generalization loss}
\begin{align}
    \metaGenLoss=\mathbbm E_{\mathrm P_{\taskRV} \mathrm P_{\bm Z^{\lenZm} |\taskRV}} \bigl[
    \mathbbm E_{\mathrm P_{W|\bm Z^{\lenZm} ,U=u}}
 [\mathrm L_{\mathrm P_{Z|\taskRV}}(W)] \bigr],
    \label{sharu_eq12}
\end{align}
where  the expectation is taken over an independently generated meta-test task $\taskRV \sim \mathrm P_\taskRV$, over the associated data set $\bm Z^\lenZm \sim \mathrm P_{\bm Z^\lenZm|\taskRV}$, and over the output of the base-learner. 
Since $\mathrm P_\taskRV$ and $\{\mathrm P_{Z|\taskRV=\task}\}_{\task \in \mathcal{T}}$ are unknown, the meta-generalization loss~\eqref{sharu_eq12}
cannot be computed.
Instead, the meta-learner can evaluate the \textit{meta-training loss}, which for a
given hyperparameter $u$, is defined as the average training loss on the meta-training set
\begin{align}
\mathrm L_{{\bm Z_{1:\lenDataSet}^\lenZm}}(u)  =\frac{1}{\lenDataSet}\sum_{i=1}^{\lenDataSet}\mathbbm E_{\mathrm P_{W|\bm Z^{\lenZm}_i ,U=u}} [\mathrm L_{\bm Z^{\lenZm}_i}(W)].
\label{sharu_eq11}
\end{align}
Here, the average is taken over the output of the base-learner.
The difference between the meta-generalization loss and the meta-training loss is the \textit{meta-generalization gap}
\begin{align}
\Delta\mathrm L(u|\bm Z_{1:\lenDataSet}^\lenZm)=\mathrm L(u)-\mathrm L_{{\bm Z_{1:\lenDataSet}^\lenZm}}(u).
\label{2020.06.24_22.18}
\end{align}
Intuitively, if the meta-generalization gap is small, on average or with high probability, then the performance of the inferred hyperparameter $u$ on the meta-training set can be taken as a reliable measure of the meta-generalization loss~\eqref{sharu_eq12}.

In this paper, we consider a stochastic meta-learner described by the conditional probability distribution $\mathrm P_{U|\bm Z_{1:\lenDataSet}^\lenZm }$, which maps the meta-training set $\bm Z_{1:\lenDataSet}^\lenZm$ to the space of hyperparameters $\mathcal{U}$.
We seek an information-theoretic upper bound on the average meta-generalization gap
\begin{align}  
\Delta \mathrm L^{\textup{avg}}=\mathbbm E_{\mathrm P_{U\bm Z^\lenZm_{1:\lenDataSet}}}  \left[\Delta\mathrm L (U|\bm Z_{1:\lenDataSet}^\lenZm)\right], \label{eq:avgmetagengap} 
\end{align}
where $\mathrm P_{U\bm Z^\lenZm_{1:\lenDataSet}}=\mathrm P_{\bm Z^\lenZm_{1:\lenDataSet}}\mathrm P_{U|\bm Z^\lenZm_{1:\lenDataSet}}$ is the joint distribution of meta-training data and $U$ induced by the meta-learner. 

\section{A CMI-Based Framework for Meta-Learning}\label{sec:CMI-framework}
In this section, we introduce the CMI framework for meta-learning. 
%
We start by noting that the average meta-generalization gap~\eqref{eq:avgmetagengap} can be decomposed as~\cite{sharu}
\begin{align}
  \Delta \mathrm L^{\textup{avg}}=\mathbbm E_{\mathrm P_{U\bm Z^\lenZm_{1:\lenDataSet}}}\biggl[ \metaGenLossRV-\mathbbm E_{\mathrm P_{\taskRV \bm Z^\lenZm}}[\Lhat_{\bm Z^{\lenZm}}(U)]\biggr] 
 \hspace{2em}\nonumber \\ +\mathbbm  E_{\mathrm P_{U\bm Z^\lenZm_{1:\lenDataSet}}}\biggl[\mathbbm E_{\mathrm P_{\taskRV \bm Z^\lenZm}}[\Lhat_{\bm Z^{\lenZm}}(U)]-\mathrm L_{{\bm Z_{1:\lenDataSet}^\lenZm}}(U)\biggr], \label{eq:decomposition}
  \end{align}
where $\Lhat_{\bm Z^{\lenZm}}(u)$ denotes the average per-task training loss:
\begin{align}
    \Lhat_{\bm Z^{\lenZm}}(u)=\mathbbm E_{\mathrm P_{W|\bm Z^{\lenZm} ,U=u}}\left[\mathrm L_{\bm Z^{\lenZm}}(W)\right].\label{L_hat}
\end{align} 
This quantity differs from the meta-training loss in that the hyperparameter $U$ does not depend on the per-task training sets  $\bm Z^{\lenZm}$ used to evaluate the loss.
On the contrary, the meta-training loss is evaluated on the meta-training data used to determine $U$. 
It also differs from the meta-generalization loss, since it averages training losses and not test losses. 
The first expectation in~\eqref{eq:decomposition} captures the average \textit{within-task generalization gap} associated to meta-test task $\taskRV \sim \mathrm P_\taskRV$, caused by the fact that only $\lenZm$ training samples per the task are available. 
The second expectation captures the average \textit{environment-level generalization gap},  caused by the fact that the meta-learner observes only $\lenDataSet$ tasks.

To obtain an upper bound on the average meta-generalization gap, we bound these two expectations separately using the CMI approach introduced in the next section. 
\subsection{Per-Task Supersample and Meta-Supersamples}
\label{sec-super-sample-single-task}
In this section, we define the per-task supersample following~\cite{Steinke2020} and we introduce the concept of meta-supersample. 

For a given task $\task_i$, we define the per-task supersample as the collection $\tilde{\bm Z}^{2\lenZm}_i=( Z^1_i,\dots, Z^{2\lenZm}_i)$ of $2\lenZm$ samples draw independently from $\mathrm P_{Z|\taskRV=\task_i}$. Let
$\bm S_i=(S_{i}^1,\dots,S_{i}^\lenZm)$  be an $\lenZm$-dimensional random vector whose elements are drawn independently from a Bernoulli distribution with parameter $0.5$, independent of  $\tilde{\bm Z}^{2\lenZm}_i$.  
We use the vector $\bm S_i$ to partition the per-task supersample $\tilde{\bm Z}^{2\lenZm}_i$ into a set of $\lenZm$ input training samples fed to the base-learner, and a set of $\lenZm$ test samples.
The vector of $\lenZm$ input training samples, which we denote as $\tilde{\bm Z}_i^{2\lenZm}(\bm S_i)$  is obtained as  $\tilde{\bm Z}_i^{2\lenZm}(\bm S_i)=(Z_i^{1+\lenZm S_{i}^1},Z_i^{
2+\lenZm S_{i}^2}, \dots, Z_i^{\lenZm+\lenZm S_{i}^\lenZm} )$.
Let  $\bar{\bm S}_i$ denotes the vector whose entries are the modulo-2 complement of the entries of $\bm S_i$.
Then $\tilde{\bm Z}_i^{2\lenZm}(\bar{\bm S}_i)$ stands for the vector of test data for task $\tau_i$.
We shall use the training sets $\tilde{\bm Z}_i^{2\lenZm}({\bm S}_i)$ to train the base-learner, while the within-task generalization loss will be evaluated on the test data $\tilde{\bm Z}^{2\lenZm}_i(\bar{\bm S}_i) $.

We now describe the construction of the \textit{meta-supersample}.  
We start by sampling $2\lenDataSet$ tasks, $\taskRV_1,\hdots,\taskRV_{2\lenDataSet}$, independently from $\mathrm P_\taskRV$. 
For each $\taskRV_i$, we generate a per-task supersample $\tilde{\bm Z}^{2\lenZm}_i$ as detailed above.
The meta-supersample is then defined as $\supSamp=( \tilde{\bm Z}^{2\lenZm}_1,\dots, \tilde{\bm Z}^{2\lenZm}_{2\lenDataSet})$.
Let now $\bm R=(R_1,\dots,R_{\lenDataSet})$ be 
a $\lenDataSet$-dimensional random vector whose elements are drawn independently according to a Bernoulli distribution with parameter $0.5$, independent of  $\supSamp$. 
We use the random vector~$\bm R$  to partition the meta-supersample into $\lenDataSet$ meta-training task datasets, and $\lenDataSet$ meta-test datasets.
Specifically, the meta-training task datasets $\supSamTask=(\bmMT_1(R_1),\dots,\bmMT_{\lenDataSet}(R_{\lenDataSet}))$ are obtained by setting  $\bmMT_i( R_i)=\tilde{\bm Z}^{2\lenZm}_{i+R_i\lenDataSet}$, $i=1,\dots,\lenDataSet$.
Finally, the meta-training data fed to the meta-learner is $\bmM(\bm R,\bm S_{1:\lenDataSet})=(\bmMT_1(R_1,\bm S_1),\dots,\bmMT_{\lenDataSet}(R_{\lenDataSet},\bm S_{\lenDataSet}))$, where $\bmMT_i( R_i,\bm S_i)=\tilde{\bm Z}^{2\lenZm}_{i+R_i\lenDataSet}(\bm S_i)$.
  As before, we let $\bar{\bm R}$ be the vector whose entries are the modulo-2 complement of the entries of $\bm R$, and use $\bmM(\bar {\bm R})$ to denote all elements of  $\supSamp$  that are not
in $\bmM({\bm R})$.

To sum up, the meta-training set $\bmM(\bm R,\bm S_{1:\lenDataSet})$ is generated by first choosing $\lenDataSet$ tasks from the meta-supersample $\supSamp$ according to $\bm R$, and then choosing $\lenZm$ samples per task from $\tilde{\bm Z}^{2\lenZm}_i$ according to  $\bm S_i$, for $i=1,\hdots,\lenDataSet$. Fig. \ref{fig:my_label} shows an example of meta supersample.

\begin{figure}[t!]
    \centering
    $ \begin{array}{c}
     \rowcolor{red!20}
    \mathrm P_{\bm Z^4|T=\tau_1}\\\mathrm P_{\bm Z^4|T=\tau_2}\\\mathrm P_{\bm Z^4|T=\tau_3}\\
     \rowcolor{red!20}
     \mathrm P_{\bm Z^4|T=\tau_4}
    \end{array}$
  $\underbrace{  \left(\begin{array}{cccc}
    \rowcolor{red!20}
   \smZ_1^1&  \smZ_1^2&   \smZ_1^3&    \smZ_1^4
  \\
\smZ_2^1&  \smZ_2^2&   \smZ_2^3&  \smZ_2^4 
\\
\smZ_3^1&  \smZ_3^2&   \smZ_3^3&  \smZ_3^4
\\
\rowcolor{red!20}
\smZ_4^1&  \smZ_4^2&     \smZ_4^3&   \smZ_4^4 
\\
  \end{array}\right)}_{\supSamp}$,
    $\underbrace{  \left(\begin{array}{cccc}
 \y \smZ_1^1&  \smZ_1^2&   \smZ_1^3&   \y \smZ_1^4 \\
\smZ_4^1&  \smZ_4^2&    \y \smZ_4^3&  \y \smZ_4^4 \\
  \end{array}\right)}_{\bmM(\bm r)}$,
    \caption{ An example of meta-supersample $\supSamp$ and $\bmM(\bm R)$ for the case $\lenZm=2$ and $\lenDataSet=2$, $\bm r=(0,1)$,
    $\bm s_1=(0,1)$ and $\bm s_2=(1,1)$.
    The rows of the meta-supersample matrix $\supSamp$ contain data samples from four tasks. 
    The shaded rows---the first and the fourth---are selected by $\bm r=(0,1)$ to form the meta-training tasks. 
    Then, the vectors $\bm s_1=(0,1)$ and $\bm s_2=(1,1)$ select the highlighted elements in $\bmM(\bm r)$
    to obtain the meta-training set  $\tilde{\bm Z}^{2\lenZm}_{1:2\lenDataSet}(\bm r,\bm s_{1:\lenDataSet})=\{\{Z_1^1,Z_1^4\},\{Z_4^3,Z_4^4\}\}$.}
    \label{fig:my_label}
    \vspace{-0.4cm}
\end{figure}


\section{CMI-Based Bounds on 
 $\Delta \mathrm L^{\textup{avg}}$
}\label{bnd-ave-meta}
In this section, we introduce a CMI-based bound on the average meta-generalization gap $\Delta \mathrm L^{\textup{avg}}$ in~\eqref{eq:avgmetagengap}. Throughout, we assume that the loss function $\ell(\cdot,\cdot)$ is bounded on the interval $[a,b]$. 
As discussed in Section~\ref{sec:CMI-framework}, to obtain an upper bound on the average meta-generalization gap, we upper-bound separately the within-task and the environment-level generalization gaps.
Towards this goal, we leverage the exponential-inequality-based approach introduced in~\cite{fredrik}.

\subsection{Main Result}
%
\begin{theorem}
\label{cor_ave_meta}
Under the assumption that the loss function is bounded as $0\leq a \leq \ell(\cdot,\cdot) \leq b < \infty$, 
the following bound on the average meta-generalization  gap~\eqref{eq:avgmetagengap} holds:
\begin{align}
\Delta \mathrm L^{\textup{avg}} &\leq\sqrt{2 (b-a)^2\left(\frac{\mathrm  I(U
      ;\bm R,\bm S_{1:\lenDataSet}|\supSamp) }{\lenDataSet}\right)}\hspace{0em}\nonumber
      \end{align}
      \begin{align}
  &+\frac{1}{\lenDataSet}\sum_{i=1}^{\lenDataSet}  \sqrt{2 (b-a)^2\left(\frac{\mathrm I(W;\bm S_i|\supSamp, R_i)}{\lenZm}\right)
               } \label{cor_ave_meta_equation}.
\end{align}
\end{theorem}

The theorem is proved using the exponential inequalities that we report in the next subsection and 
Appendix \ref{IEEEproof_cor_ave_meta}. 
The first term in~\eqref{cor_ave_meta_equation} accounts for the environment-level generalization gap via the CMI $\mathrm I(U;\bm R,\bm S_{1:\lenDataSet}|\supSamp)$ divided by the number $\lenDataSet$ of meta-training input tasks. The CMI  $\mathrm I(U;\bm R,\bm S_{1:\lenDataSet}|\supSamp)$ measures the information the meta-learner output reveals about the environment-level partition $\bm R$ and per-task partition $\bm S_{1:\lenDataSet}$  of the meta-supersample $\supSamp$, when the supersample $\supSamp$ is given. 
The second term in~\eqref{cor_ave_meta_equation} accounts for the within-task generalization gap through the CMI $\mathrm I(W;\bm S_i|\supSamp,R_i)$, 
or equivalently
$\mathrm I(W;\bm S_i|\tilde{\bm Z}^{2\lenZm}_{i+R_i\lenDataSet})$.
Consistent with the per-task generalization gap bounds of \cite{Steinke2020}, this second term measures the information the base-learner output reveals about the partitioning $\bm S_i$ of the per-task supersample $\tilde{\bm Z}^{2\lenZm}_{i+R_i\lenDataSet}$. 

The CMI-based bound~\eqref{cor_ave_meta_equation} recovers the CMI bound for conventional learning in~\cite[ Thm. 2]{Steinke2020} as special case. For conventional learning, the hyperparameter $u$  is fixed \textit{a priori}. Moreover, the 
 task environment distribution can be assumed to be a delta function centered at some task $\task \in \mathcal{T}$, and the meta-training set with $\lenDataSet=1$ reduces to the training set of task $\task$. Hence, the first MI in \eqref{cor_ave_meta_equation} is zero, and~\eqref{cor_ave_meta_equation} reduces to 
\begin{align}
  \hspace{-0.3cm}\mathbbm E_{\mathrm P_{W\bm Z^\lenZm}} \big[\Delta\mathrm L(W|\bm Z^\lenZm)\big] \leq  \sqrt{\frac{ 2(b-a)^2 \mathrm I(W
      ;\bm S|\tilde{\bm Z}^{2\lenZm}) }{M}}, 
\end{align} which recovers the bound in~\cite[ Thm. 2]{Steinke2020}.
{
It can be verified that
the CMI-based bound~\eqref{cor_ave_meta_equation} is  bounded by $\sqrt{2(b-a)^2\log2}(\sqrt{1+\lenZm}+1)$.  
A comparison with other information-theoretic bounds is presented in Section \ref{example}.
}
\subsection{Exponential Inequalities}
The proof of Theorem~\ref{cor_ave_meta} relies on exponential inequalities that are used to bound the two terms in the decomposition~\eqref{eq:decomposition}.

We start by expressing the within-task generalization gap, \textit{i.e.}, 
the first expectation in the decomposition~\eqref{eq:decomposition}, in the following equivalent form, which makes explicit its dependence on the per-task supersamples and meta-supersample,
\begin{align}
   &\mathbbm E_{\mathrm P_{U\bm Z^\lenZm_{1:\lenDataSet}}}\left[ \metaGenLossRV-\mathbbm E_{\mathrm P_{\taskRV \bm Z^\lenZm}}[\Lhat_{\bm Z^{\lenZm}}(U)]\right]\nonumber\\
   & =\mathbbm E_{\mathrm P_{U\bm Z^\lenZm_{1:\lenDataSet}}}
   \mathbbm E_{\mathrm P_{\taskRV\bm Z^\lenZm  } } 
   \mathbbm E_{\mathrm P_{W|\bm Z^\lenZm U}}\left[
 \mathrm L_{\mathrm P_{Z|\taskRV}}(W)-  \mathrm L_{\bm Z^{\lenZm}}(W)
  \right]\label{2020.10.18_14.54a}\\
  & =\mathbbm E_{\mathrm P_{U\supSamp\bm R\bm S_{1:\lenDataSet}}}\biggl[ \frac{1}{\lenDataSet}\sum_{i=1}^\lenDataSet
  \mathbbm E_{\mathrm P_{W|\bmMT_i( \bar R_i,\bm S_i), U}}\Bigl[
\mathrm L_{\bmMT_i( \bar R_i,\bar{\bm S}_i)}(W)\nonumber\\
&\hspace{13em}- \mathrm L_{\bmMT_i( \bar R_i,\bm S_i)}(W)
  \Bigr]\biggr]
  \label{2020.10.18_14.54b} 
  \\
  & = \frac{1}{\lenDataSet}\sum_{i=1}^\lenDataSet
   \mathbbm E_{\mathrm P_{\supSamp {R}_i \bm S_i}\mathrm P_{W|\bmMT_i( \bar R_i,\bm S_i)}}\Bigl[
\mathrm L_{\bmMT_i( \bar R_i,\bar{\bm S}_i)}(W)\nonumber\\
&\hspace{13.5em}- \mathrm L_{\bmMT_i( \bar R_i,\bm S_i)}(W)
  \Bigr].
  \label{2020.10.18_14.54d}
\end{align}
The first equality~\eqref{2020.10.18_14.54b} holds since the meta-learner is trained on the meta-supersample $\bmM(\bm R, \bm S_{1:\lenDataSet})$, and since the average over $\mathrm P_{\taskRV\bm Z^\lenZm  } $ in \eqref{2020.10.18_14.54a} can be evaluated on the independent test data set $\bmM(\bar{\bm R},\bm S_{1:\lenDataSet})$. Note that the sets $\bmM (\bm R, \bm S_{1:\lenDataSet})$ and $\bmM (\bar{\bm R}, \bm S_{1:\lenDataSet})$ represent independent datasets from two different environment-level tasks, even if they share the same $\bm S_{1:\lenDataSet}$. Finally, we obtain~\eqref{2020.10.18_14.54d} by taking the expectation inside the sum and by marginalizing over $U$.

To keep notation compact, let the term within the average in~\eqref{2020.10.18_14.54d} be defined as 
\begin{align}
  \genL(W,\supSamp, {R}_i,\bm S_i)=  \mathrm L_{\bmMT_i(R_i,\bar{\bm S}_i)}(W)-
    \mathrm L_{\bmMT_i(R_i,\bm S_i)}(W).\label{genl}
\end{align}
We now present a task-level exponential inequality that will be useful to bound the expectation of this term. For generality, the bound is expressed in terms of the Radon–Nikodym derivatives of the relevant distributions (see, e.g., \cite[Sec. 17.1]{poly_lec}).
\begin{proposition}
\label{exp-ineq-super-sample-single-task}
For all $\lambda \in \mathbbm{R}$, 
we have
\begin{multline}
        \mathbbm E_{\mathrm P_{W\supSamp\bm S_i R_i}}\bigg[
    \exp\Bigl(
       \lambda\genL(W,\supSamp, {R}_i,\bm S_i)\hspace{2em}\nonumber\\-\frac{\lambda^2(b-a)^2}{2\lenZm}
       - \imath(W;\bm S_i|\supSamp, R_i)
    \Big)
    \bigg]\leq 1,
        \label{prop-gen-hat-single}
\end{multline} 
where     $ \imath(W;\bm S|\supSamp, R)$ is the \textit{conditional information density}
\begin{align}
     \imath(W;\bm S|\supSamp, R)
    =\log\frac{ \mathrm P_{W\bm S|\supSamp R}}{\mathrm P_{W|\supSamp  R }  \mathrm P_{\bm S}}.
\end{align}
\end{proposition}
\begin{IEEEproof}
See Appendix \ref{IEEEproof_exp-ineq-super-sample-single-task}.
\end{IEEEproof}

We now derive a similar exponential inequality to bound the average environment-level generalization gap, i.e., the second expectation on the right-hand side of~\eqref{eq:decomposition}. 
Let 
\begin{align}
  \genLU(U,\supSamp,\bm R,\bm S_{1:\lenDataSet})\hspace{13em}\nonumber\\=\mathrm  L_{\bmM(\bar{\bm R},\bar{\bm S}_{1:\lenDataSet})}(U) 
 -
    \mathrm  L_{\bmM({\bm R},\bm S_{1:\lenDataSet})}(U),\label{gen_hat_for_meta}
\end{align}
where $\mathrm L_{{\bm Z_{1:\lenDataSet}^\lenZm}}(u)$ was defined in~\eqref{sharu_eq11}. 
Using~\eqref{gen_hat_for_meta}, and following steps similar to the ones leading to~\eqref{2020.10.18_14.54d}, we can express the environment-level generalization gap as
\begin{align}
    &\mathbbm E_{\mathrm P_{U\bm Z^\lenZm_{1:\lenDataSet}}}\Big[\mathbbm E_{\mathrm P_{\taskRV, \bm Z^\lenZm}}\bigl[\Lhat_{\bm Z^{\lenZm}}(U)\bigr]-\frac{1}{\lenDataSet}\sum_{i=1}^{\lenDataSet}\Lhat_{\bm Z_i^\lenZm}(U)
  \Big]=\nonumber
      \\
    & 
      \mathbbm E_{\mathrm P_{\supSamp\bm R\bm S_{1:\lenDataSet}}\mathrm P_{U|\supSamp\bm R\bm S_{1:\lenDataSet}
      }}\left[
          \genLU(U,\supSamp,\bm R,\bm S_{1:\lenDataSet})\right]. \label{2020.06.24_22.02}
\end{align}
The next proposition states the relevant inequality.
The proof is similar to that of Proposition~\ref{exp-ineq-super-sample-single-task} and, hence, omitted.
\begin{proposition}
\label{exp-ineq-super-sample-multi-task}
For all $\lambda \in \mathbbm{R}$, the following exponential inequality holds
\begin{align}
     &\mathbbm E_{\mathrm P_{U\supSamp
     \bm R \bm S_{1:\lenDataSet}}}\bigg[\exp
    \Bigl(\lambda\genLU(U,\supSamp,\bm R,\bm S_{1:\lenDataSet})
    \nonumber\\&\hspace{2em}-   \frac{\lambda^2(b-a)^2}{2\lenDataSet}
   - \imath\bigl(
    U
    ;\bm R,\bm S_{1:\lenDataSet}|\supSamp
    \bigr)
    \Big)
    \bigg]\leq 
1,
      \label{prop-gen-hat-multi}
\end{align}
where $\mathrm P_{U\supSamp
     \bm R \bm S_{1:\lenDataSet}}=\mathrm P_{\supSamp
     \bm R \bm S_{1:\lenDataSet}}\mathrm P_{U|\supSamp
     \bm R \bm S_{1:\lenDataSet}}$ and $ \imath(
    U
    ;\bm R,\bm S_{1:\lenDataSet}|\supSamp
    )$ is the  {conditional information density}, 
\begin{align}
     \imath\bigl(
    U
    ;\bm R,\bm S_{1:\lenDataSet}|\supSamp
    \bigr)=\log\frac{\mathrm P_{U
     \bm R \bm S_{1:\lenDataSet}|\supSamp}}{
    \mathrm P_{U|\supSamp
 }
    \mathrm P_{
     \bm R \bm S_{1:\lenDataSet}}}.
\end{align}
\end{proposition}

Using these inequalities, the proof of Theorem \ref{cor_ave_meta} can be completed as detailed in Appendix \ref{IEEEproof_cor_ave_meta}.

\section{Example}
\label{example}
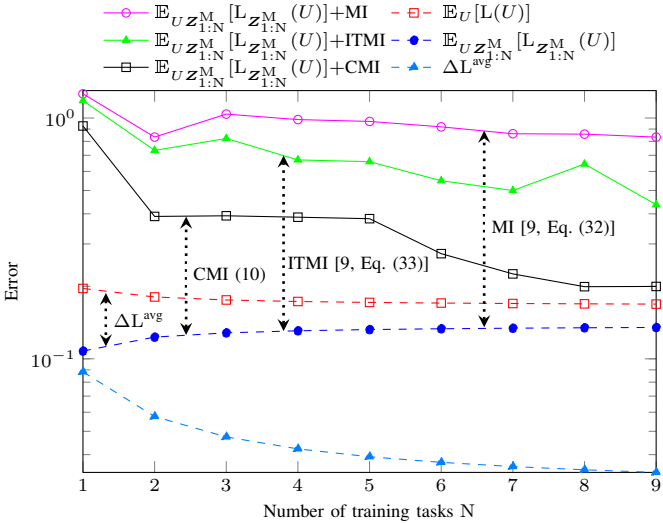
\begin{figure}[!t]
\centering
%
%

\definecolor{mycolor1}{rgb}{1.00000,0.00000,1.00000}%
\definecolor{mycolor2}{rgb}{0.00000,0.50000,2.50000}%
\definecolor{mycolor3}{rgb}{.5,.5,.5}%
\begin{tikzpicture}

\begin{axis}[%
width=3in,
height=2in,
at={(0.758in,0.481in)},
scale only axis,
xmin=1,
xmax=9,
ymin=0.0337499120195,
ymax=1.3,
ymode=log,
yminorticks=true,
xlabel=Number of training tasks $\lenDataSet$,
ylabel= Error,
axis background/.style={fill=white},
legend style={at={(.02,1.01)}, anchor=south west, legend cell align=left, align=left, draw=none, legend columns=2}
]
\addplot [color=mycolor1,mark=o,mark options={scale=.8}]
  table[row sep=crcr]{%
1	1.26082922687441\\
2	0.833673258675518\\
3	1.03836840497561\\
4	0.985379300818924\\
5	0.967999876433366\\
6	0.919169737860867\\
7	0.860515951114028\\
8	0.857284809209278\\
9	0.833303395875522\\
};
\addlegendentry{$\mathbbm E_{ U{\bm Z_{1:\lenDataSet}^\lenZm}}[\mathrm L_{{\bm Z_{1:\lenDataSet}^\lenZm}}(U)]+$MI}

\addplot [color=red,dashed,mark=square, mark options={solid,scale=0.8}]
  table[row sep=crcr]{%
1	0.1960046432955\\
2	0.180673658996625\\
3	0.175563330897\\
4	0.173008166847188\\
5	0.1714750684173\\
6	0.170453002797375\\
7	0.169722955926\\
8	0.169175420772469\\
9	0.1687495600975\\
};
\addlegendentry{$ \mathbbm E_{U}[\mathrm L(U)]$ 
}

\addplot [color=green,mark=triangle*,mark options={scale=.8}]
  table[row sep=crcr]{%
1	1.18095460559877\\
2	0.733813145978727\\
3	0.822386383971202\\
4	0.669991397909248\\
5	0.658632435553223\\
6	0.549131147986061\\
7	0.499957310329753\\
8	0.644139664027929\\
9	0.436822440281136\\
};
\addlegendentry{$\mathbbm E_{ U{\bm Z_{1:\lenDataSet}^\lenZm}}[\mathrm L_{{\bm Z_{1:\lenDataSet}^\lenZm}}(U)]+$ITMI}

\addplot [color=blue,dashed,mark=*,mark options={scale=.8}]
  table[row sep=crcr]{%
1	0.10774456488\\
2	0.123075549178875\\
3	0.1281858772785\\
4	0.130741041328312\\
5	0.1322741397582\\
6	0.133296205378125\\
7	0.1340262522495\\
8	0.134573787403031\\
9	0.134999648078\\
};
\addlegendentry{$\mathbbm E_{ U{\bm Z_{1:\lenDataSet}^\lenZm}}[\mathrm L_{{\bm Z_{1:\lenDataSet}^\lenZm}}(U)]$
}

\addplot [color=black,mark=square,mark options={scale=.8}]
  table[row sep=crcr]{%
1	0.926767706517609\\
2	0.390303673755108\\
3	0.3926858772785\\
4	0.387441041328312\\
5	0.381926791724986\\
6	0.273352648443265\\
7	0.225259948250049\\
8	0.199378896544822\\
9	0.199801993046326\\
};
\addlegendentry{$\mathbbm E_{ U{\bm Z_{1:\lenDataSet}^\lenZm}}[\mathrm L_{{\bm Z_{1:\lenDataSet}^\lenZm}}(U)]+$CMI}

\addplot [color=mycolor2,dashed,mark=triangle*]
  table[row sep=crcr]{%
1	0.0882600784155\\
2	0.05759810981775\\
3	0.0473774536185\\
4	0.042267125518875\\
5	0.0392009286591\\
6	0.03715679741925\\
7	0.0356967036765\\
8	0.0346016333694375\\
9	0.0337499120195\\
};
\addlegendentry{$\Delta \mathrm L^{\textup{avg}}
$ 
}

\begin{scope}[
   line width=1pt,
   myarrowlabel/.style={anchor=west,fill=white}
   ]
\draw[<->,dotted]  (rel axis cs:\xA, .47) --node[myarrowlabel]
{$\Delta \mathrm L^{\textup{avg}}$} (rel axis cs:\xA,0.33);
\draw[<->, dotted] (rel axis cs:\xB,0.36) --node[myarrowlabel]
{CMI~\eqref{cor_ave_meta_equation}} (rel axis cs:\xB, .67);
\draw[<->, dotted] (rel axis cs:\xC,0.37) --node[right of=myarrowlabel, above=-0.5cm]
{ITMI~\cite[Eq. (33)]{sharu}} (rel axis cs:\xC, .83);
\draw[<->, dotted] (rel axis cs:\xD,0.38) --node[myarrowlabel]
{MI~\cite[Eq. (32)]{sharu}} (rel axis cs:\xD, .895);

\end{scope}
\end{axis}
\end{tikzpicture}%
\caption{
Comparison between the CMI-based bound~\eqref{cor_ave_meta_equation} and both the MI-based and the ITMI-based bounds reported in~\cite[Eq.~(32)]{sharu} and~\cite[Eq.~(33)]{sharu}, respectively. In the figure, $|\mathcal T|=10$ with $\mathrm P_\taskRV=[0.05,0.1,0.02,0.2 ,0.01, 0.05, 0.02, 0.15, 0.1, 0.3 ]$,
$\lenZm=5$, and $\alpha=0.5$.} 
\label{fig_sup2}
\end{figure}

{
We  consider the example of mean estimation of a Bernoulli process studied in \cite{sharu}. 
Specifically, we assume a finite set of tasks
$\mathcal T$ 
and a task distribution $\mathrm P_{\taskRV}$.
For a given task $\task_i \in\mathcal T$, we assume that the data follow a Bernoulli distribution with mean $\mu_{\task_i}$.
We also adopt the loss function $\ell(w,z)=(w-z)^2$. 
The base-learner's output $W_i$ for task $\tau_i$ is chosen (deterministically) as a  convex combination of the sample average $D_i=\frac{1}{\lenZm}\sum_{j=1}^\lenZm Z_{i}^j$, and of a hyperparameter  vector $u$. 
Specifically, we set $W_i=\alpha D_i+(1-\alpha)u$  with some $\alpha\in[0,1]$.
The hyperparameter $u$ is chosen (deterministically) so as to minimize the resulting
meta-training empirical loss function
$
  \mathrm L_{{\bm Z_{1:\lenDataSet}^\lenZm}}(u)  =\frac{1}{\lenDataSet}\sum_{i=1}^{\lenDataSet}\frac{1}{\lenZm}\sum_{j=1}^{\lenZm}(W-Z_i^j)^2$,
 yielding  $U=\frac{1}{\lenDataSet}\sum_{i=1}^{\lenDataSet}D_i$ \cite{sharu}.
 
In Fig.~\ref{fig_sup2}, we compare the CMI-based bound~\eqref{cor_ave_meta_equation} with both the
MI-based and the individual task mutual information (ITMI)-based  bounds 
reported in~\cite[Eq.~(32)]{sharu} and in~\cite[Eq.~(33)]{sharu}, respectively.
We assume $|\mathcal T|=10$, $\mathrm P_\taskRV=[0.05,0.1,0.02,0.2 ,0.01, 0.05, 0.02, 0.15, 0.1, 0.3 ]$,  $\alpha=0.5$, and $\lenZm=5$.
The bounds are plotted as a function of the number $\lenDataSet$ of available training tasks. 
As shown in the figure, for this example the CMI-based bound provided a more accurate prediction of the average meta-generalization loss $\mathbbm{E}[\mathrm L(U)]$, which can be computed in closed-form for this example, \cite{sharu}, than the MI-based and the ITMI-based bounds.}
\section{Conclusion}
In this paper, we have derived a CMI-based bound on the average meta-generalization gap, which was demonstrated via  an example to potentially result in a more accurate estimate of the meta-generalization performance as compared to previously derived ITMI- and MI-based bounds.
 Adapting the CMI-based bound~\eqref{cor_ave_meta_equation} to the PAC-Bayesian setting, along the lines of~\cite{hellstrom20-12a}, may result in data-dependent high-probability bounds that are nonvacuous when applied to models such as deep neural networks. We leave this aspect to future work.
\appendices
 
\section{Proof of Proposition \ref{exp-ineq-super-sample-single-task}}
\label{IEEEproof_exp-ineq-super-sample-single-task}
Since $\ell(w,z)$ is bounded, 
 $\genL(w,\supsamp, {r}_i,\bm S_i)$ is   bounded on $[a-b,b-a]$, and hence, 
$\genL(w,\supsamp, {r}_i,\bm S_i)$ is subgaussian with parameter $(b-a)/\sqrt{\lenZm}$  \cite[Ex.~2.4]{Wainwright}. 
Furthermore, since $
  \mathbbm E_{\mathrm P_{\bar{\bm S}_i}} [\mathrm L_{\bmMT_i( r_i,\bar{\bm S}_i)}(W)]=
    \mathbbm E_{\mathrm P_{\bm S_i}}[\mathrm L_{\bmMT_i( r_i,{\bm S}_i)}(W)] 
$, 
we have   $\mathbbm E_{\mathrm P_{\bm S_i}}
    [\genL(w,\supsamp, {r}_i,\bm S_i)]=0$,
Thus, 
for every $w$, $\supsamp$, 
${r}_i$, 
\begin{equation}
\mathbbm E_{\mathrm P_{\bm S_i}} \biggl[ \exp\bigl(
\lambda \genL(w,\supsamp, {r}_i,\bm S_i)
\bigr)\biggr]\leq \exp\left(\frac{\lambda^2(b-a)^2}{2\lenZm}
    \right)
    .
\label{2020.07.04_13.06}
\end{equation}
Taking an additional expectation over $\mathrm P_{W\supSamp R_i}$, we find that 
\begin{align}
   \mathbbm E_{\mathrm P_{\bm S_i}\mathrm P_{W\supSamp R_i}}\biggr[   \exp\biggl(
\lambda\Big( \genL(W,\supSamp, {R}_i,\bm S_i)\big)\hspace{1em}\nonumber\\-\frac{\lambda^2(b-a)^2}{2\lenZm}
    \bigg)\biggr]\leq  1.
\label{2020.06.24_10.56}
\end{align}
 The desired result then follows from a change of measure from $\mathrm P_{\bm S_i}\mathrm P_{W\supSamp R_i}$ to $\mathrm P_{W\supSamp\bm S_i R_i}$ \cite[Prop. 17.1(4)]{poly_lec}. 

 \section{Proof of Theorem \ref{cor_ave_meta}}
\label{IEEEproof_cor_ave_meta}
Substituting~\eqref{genl} into~\eqref{2020.10.18_14.54d} and then~\eqref{genl} and~\eqref{2020.06.24_22.02} into~\eqref{eq:decomposition}, 
\begin{align}
 & \mathbbm E_{\mathrm P_{U\bm Z^\lenZm_{1:\lenDataSet}}}  \left[\Delta\mathrm L (U|\bm Z^\lenZm_{1:\lenDataSet})\right]\nonumber \\
      &=\frac{1}{\lenDataSet}\sum_{i=1}^\lenDataSet
\mathbbm E_{\mathrm P_{W\supSamp\bm S_i R_i}}\left[\genL(W,\supSamp, R_i,\bm S_i)\right] \nonumber \\
& + \mathbbm E_{\mathrm P_{\supSamp\bm R\bm S_{1:\lenDataSet}}\mathrm P_{U|\supSamp\bm R\bm S_{1:\lenDataSet}
      }}\left[
          \genLU(U,\supSamp,\bm R,\bm S_{1:\lenDataSet})\right]. \label{eq:decomposition_in_cmi_form}
\end{align}

To bound the two terms on the right-hand side of~\eqref{eq:decomposition_in_cmi_form}, we use the exponential inequalities in Proposition \ref{exp-ineq-super-sample-single-task} and~\ref{exp-ineq-super-sample-multi-task}.
Specifically, applying Jensen’s inequality to~\eqref{prop-gen-hat-single} and~\eqref{prop-gen-hat-multi},
we find the following inequalities
\begin{align}
    &\exp\biggl(
        \lambda\mathbbm E_{\mathrm P_{W\supSamp\bm S_i  R_i}}\left[\genL(W,\supSamp, {R}_i,\bm S_i)\right]\hspace{3em}\nonumber\\&-\frac{\lambda^2(b-a)^2}{2\lenZm}
       -\mathrm I(W;\bm S_i|\supSamp,R_i)
    \bigg)
    \leq 1,
    \label{boro_baba1}
    \end{align}
    and
\begin{align}
     &\exp\biggl(\lambda \mathbbm E_{\mathrm P_{U\supSamp
     \bm R\bm S_{1:\lenDataSet}}}   \left[\genLU(U,\supSamp,\bm R,\bm S_{1:\lenDataSet})\right]\hspace{3em}\nonumber\\
  & -\frac{\lambda^2(b-a)^2}{2\lenDataSet}
    -\mathrm I(U
    ;\bm R,\bm S_{1:\lenZm}|\supSamp) 
      \bigg)\leq 1,
      \label{boro_baba2}
\end{align}
where \eqref{boro_baba1} and \eqref{boro_baba2} are valid for every $\lambda\in\mathbbm R$. Taking the $\log$ on both sides of the above inequalities, and optimizing over~$\lambda$, we obtain upper bounds on the two expectations in~\eqref{eq:decomposition_in_cmi_form} (similar to the one reported in~\cite[Cor.~5]{fredrik}), which, when substituted into~\eqref{eq:decomposition_in_cmi_form}, give the desired bound.

\bibliographystyle{IEEEtran}
\bibliography{exemple_bibtex}

\begin{thebibliography}{10}
\providecommand{\url}[1]{#1}
\csname url@samestyle\endcsname
\providecommand{\newblock}{\relax}
\providecommand{\bibinfo}[2]{#2}
\providecommand{\BIBentrySTDinterwordspacing}{\spaceskip=0pt\relax}
\providecommand{\BIBentryALTinterwordstretchfactor}{4}
\providecommand{\BIBentryALTinterwordspacing}{\spaceskip=\fontdimen2\font plus
\BIBentryALTinterwordstretchfactor\fontdimen3\font minus
  \fontdimen4\font\relax}
\providecommand{\BIBforeignlanguage}[2]{{%
\expandafter\ifx\csname l@#1\endcsname\relax
\typeout{** WARNING: IEEEtran.bst: No hyphenation pattern has been}%
\typeout{** loaded for the language `#1'. Using the pattern for}%
\typeout{** the default language instead.}%
\else
\language=\csname l@#1\endcsname
\fi
#2}}
\providecommand{\BIBdecl}{\relax}
\BIBdecl

\bibitem{schmidhuber1987evolutionary}
J.~Schmidhuber, ``Evolutionary {P}rinciples in {S}elf-{R}eferential {L}earning,
  or {O}n {L}earning {H}ow to {L}earn: {T}he {M}eta-meta-... {H}ook,'' Ph.D.
  dissertation, Technische Universit{\"a}t M{\"u}nchen, 1987.

\bibitem{thrun1998learning}
S.~Thrun and L.~Pratt, ``Learning to {L}earn: {I}ntroduction and {O}verview,''
  in \emph{Learning to Learn}.\hskip 1em plus 0.5em minus 0.4em\relax Springer,
  1998, pp. 3--17.

\bibitem{finn2017model}
C.~Finn, P.~Abbeel, and S.~Levine, ``Model-agnostic meta-learning for fast
  adaptation of deep networks,'' in \emph{Proc. Int. Conf. Machine Learning
  (ICML)}, Sydney, Australia, Aug. 2017.

\bibitem{li2017meta}
\BIBentryALTinterwordspacing
Z.~Li, F.~Zhou, F.~Chen, and H.~Li, ``{M}eta-{SGD}: Learning to learn quickly
  for few-shot learning,'' arXiv:1707.09835, 2017. [Online]. Available:
  \url{http://arxiv.org/abs/1707.09835}
\BIBentrySTDinterwordspacing

\bibitem{Xu2017}
A.~Xu and M.~Raginsky, ``Information-theoretic analysis of generalization
  capability of learning algorithms,'' in \emph{Proc. Conf. Neural Inf.
  Process. Syst. (NeurIPS)}, Long Beach, CA, {USA}, Dec. 2017.

\bibitem{Russo2016ControllingTheory}
D.~Russo and J.~Zou, ``Controlling {bias} in {adaptive} {data} {analysis}
  {using} {information} {theory},'' in \emph{Proc. Artif. Intell. Statist.
  (AISTATS)}, Cadiz, Spain, May 2016.

\bibitem{Bu}
Y.~Bu, S.~Zou, and V.~V. Veeravalli, ``Tightening {mutual} {information}
  {based} {bounds} on {generalization} {error},'' in \emph{Proc. IEEE Int.
  Symp. Inf. Theory (ISIT)}, Paris, France, July 2019.

\bibitem{Negrea2019Information-TheoreticEstimates}
J.~Negrea, M.~Haghifam, G.~K. Dziugaite, A.~Khisti, and D.~M. Roy,
  ``{Information-theoretic generalization bounds for SGLD via data-dependent
  estimates},'' in \emph{Proc. Conf. Neural Inf. Process. Syst. (NeurIPS)},
  Vancouver, Canada, Dec. 2019.

\bibitem{sharu}
\BIBentryALTinterwordspacing
S.~Theresa~Jose and O.~Simeone, ``Information-theoretic generalization bounds
  for meta-learning and applications,'' arXiv:2005.04372, 2020. [Online].
  Available: \url{http://arxiv.org/abs/2005.04372}
\BIBentrySTDinterwordspacing

\bibitem{Steinke2020}
T.~Steinke and L.~Zakynthinou, ``{Reasoning about generalization via
  conditional mutual information},'' in \emph{Proc. Conf. Learn Theory (COLT)},
  Graz, Austria, July 2020.

\bibitem{baxter}
J.~Baxter, ``A model of inductive bias learning,'' \emph{Journal of Artif.
  Intell. Research}, vol.~12, pp. 149--198, Mar. 2000.

\bibitem{pentina2014pac}
A.~Pentina and C.~Lampert, ``A {PAC}-{Bayesian} bound for lifelong learning,''
  in \emph{Proc. Int. Conf. on Machine Learning (ICML)}, Beijing, China, June
  2014.

\bibitem{rothfuss2020pacoh}
\BIBentryALTinterwordspacing
J.~Rothfuss, V.~Fortuin, and A.~Krause, ``{PACOH}: Bayes-optimal meta-learning
  with {PAC}-guarantees,'' arXiv:2002.05551, 2020. [Online]. Available:
  \url{http://arxiv.org/abs/2002.05551}
\BIBentrySTDinterwordspacing

\bibitem{amit2018meta}
R.~Amit and R.~Meir, ``Meta-learning by adjusting priors based on extended
  {PAC}-{Bayes} theory,'' in \emph{Proc. Int. Conf. on Machine Learning
  (ICML)}, Stockholm, Sweden, July 2018.

\bibitem{denevi2018incremental}
G.~Denevi, C.~Ciliberto, D.~Stamos, and M.~Pontil, ``Incremental
  learning-to-learn with statistical guarantees,'' \emph{UAI}, pp. 457--466,
  2018.

\bibitem{paper1}
G.~Denevi, C.~Ciliberto, R.~Grazzi, and M.~Pontil, ``Learning-to-learn
  stochastic gradient descent with biased regularization,'' in \emph{Proc. Int.
  Conf. on Machine Learning (ICML)}, Long Beach, CA, USA, June 2019.

\bibitem{denevi2020advantage}
G.~Denevi, M.~Pontil, and C.~Ciliberto, ``The advantage of conditional
  meta-learning for biased regularization and fine-tuning,'' in \emph{Proc.
  Conf. Neural Inf. Process. Syst. (NeurIPS)}, Long Beach, CA, {USA}, Dec.
  2020.

\bibitem{paper4}
G.~Denevi, C.~Ciliberto, D.~Stamos, and M.~Pontil, ``Learning to learn around a
  common mean,'' in \emph{Proc. Conf. Neural Inf. Process. Syst. (NeurIPS)},
  Montreal, Canada, Dec. 2018.

\bibitem{paper3}
\BIBentryALTinterwordspacing
M.~Konobeev, I.~Kuzborskij, and C.~Szepesv\'ari, ``{On Optimality of
  Meta-Learning in Fixed-Design Regression with Weighted Biased
  Regularization},'' arXiv:2011.00344, 2020. [Online]. Available:
  \url{https://arxiv.org/abs/2011.00344}
\BIBentrySTDinterwordspacing

\bibitem{paper2}
M.-F. Balcan, M.~Khodak, and A.~Talwalkar, ``Provable guarantees for
  gradient-based meta-learning,'' in \emph{Proc. Int. Conf. Machine Learning
  (ICML)}, Long Beach, CA, USA, June 2019.

\bibitem{paper5}
M.~Khodak, M.-F.~F. Balcan, and A.~S. Talwalkar, ``Adaptive gradient-based
  meta-learning methods,'' in \emph{Proc. Conf. Neural Inf. Process. Syst.
  (NeurIPS)}, Vancouver, Canada, Dec. 2019.

\bibitem{zhang16-11a}
C.~Zhang, S.~Bengio, M.~Hardt, B.~Recht, and O.~Vinyals, ``Understanding deep
  learning requires rethinking generalization,'' in \emph{Proc. Int. Conf.
  Learning Representations, (ICLR) 2017}, Toulon, France, Apr. 2017.

\bibitem{fredrik}
F.~{Hellström} and G.~{Durisi}, ``Generalization bounds via information
  density and conditional information density,'' \emph{IEEE J. Sel. Areas Inf.
  Theory}, vol.~1, no.~3, pp. 824--839, Nov. 2020.

\bibitem{poly_lec}
Y.~Polyanskiy and Y.~{Wu}, \emph{Lecture notes on Information Theory}, MIT
  (6.441), UIUC (ECE 563), Yale (STAT 664), 2019.

\bibitem{hellstrom20-12a}
\BIBentryALTinterwordspacing
F.~Hellstr{\"o}m and G.~Durisi, ``Nonvacuous loss bounds with fast rates for
  neural networks via conditional information measures,'' Dec. 2020. [Online].
  Available: \url{https://arxiv.org/abs/2010.11552}
\BIBentrySTDinterwordspacing

\bibitem{Wainwright}
M.~Wainwright, \emph{High-Dimensional Statistics: A Non-Asymptotic
  Viewpoint}.\hskip 1em plus 0.5em minus 0.4em\relax Cambridge, U.K: Cambridge
  Univ. Press, 2019.

\end{thebibliography}

\end{document}